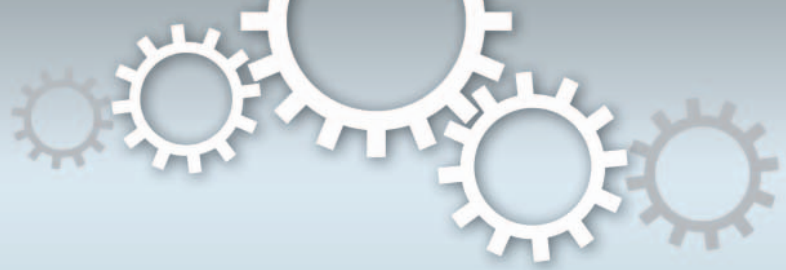

# SCIENTIFIC REPORTS

**OPEN**

# Template-Cut: A Pattern-Based Segmentation Paradigm

Jan Egger[1,2,3], Bernd Freisleben[2], Christopher Nimsky[3] & Tina Kapur[1]

[1]Department of Radiology, Brigham and Women's Hospital, Harvard Medical School, Boston, MA, USA, [2]Department of Mathematics and Computer Science, University of Marburg, Marburg, Germany, [3]Department of Neurosurgery, University of Marburg, Marburg, Germany.

We present a scale-invariant, template-based segmentation paradigm that sets up a graph and performs a graph cut to separate an object from the background. Typically graph-based schemes distribute the nodes of the graph uniformly and equidistantly on the image, and use a regularizer to bias the cut towards a particular shape. The strategy of uniform and equidistant nodes does not allow the cut to prefer more complex structures, especially when areas of the object are indistinguishable from the background. We propose a solution by introducing the concept of a "template shape" of the target object in which the nodes are sampled non-uniformly and non-equidistantly on the image. We evaluate it on 2D-images where the object's textures and backgrounds are similar, and large areas of the object have the same gray level appearance as the background. We also evaluate it in 3D on 60 brain tumor datasets for neurosurgical planning purposes.

G raph-based approaches to segmentation have gained popularity in recent years both in the general computer vision literature as well as in applied biomedical research[1–4] because of their ability to provide a globally optimal solution. This stands especially in contrast to another popular segmentation technique, deformable models[5,6], that can be easily confounded by local minima during the iterative segmentation (expansion) process. In this study, we present a novel graph-based algorithm for segmenting 2D and 3D objects. The algorithm sets up a graph and performs a graph cut to separate an object from the background. Typical graph-based segmentation algorithms distribute the nodes of the graph uniformly and equidistantly on the image. Then, a regularizer is added[7,8] to bias the cut towards a particular shape[9]. This strategy does not allow the cut to prefer more complex structures, especially when areas of the object are indistinguishable from the background. We solve this problem by introducing the concept of a "template" shape of the object when sampling the graph nodes, i.e., the nodes of the graph are distributed non-uniformly and non-equidistantly on the image. This type of template-based segmentation is particularly applicable to medical imagery, where it is easy to obtain initial landmarking[10,11] and patient orientation from the information stored in the image headers. To evaluate our method, we demonstrate results on 2D images where the gray level appearance of the objects and backgrounds are quite similar. In 3D, we demonstrate the results of the segmentation algorithm on 60 clinical Magnetic Resonance Imaging (MRI) datasets of brain tumor (glioblastoma multiforme and pituitary adenoma) patients to support the time-consuming manual slice-by-slice segmentation process typically performed by neurosurgeons.

Evolving from the cerebral supportive cells, gliomas are the most common primary brain tumors. The grading system for astrocytomas according to the *World Health Organization (WHO)* subdivides grades I–IV, where grade I tumors tend to be least aggressive[12]. Approximately 70% of the diagnosed tumors are malignant gliomas (anaplastic astrocytoma *WHO* grade III, glioblastoma multiforme *WHO* grade IV). Subject to its histopathological appearance, the grade IV tumor is given the name glioblastoma multiforme (GBM). The GBM is the most frequent malignant primary tumor and is one of the most malignant human neoplasms. Due to their biological behavior, surgery alone cannot cure this disease. Thus, current interdisciplinary therapeutic management combines maximum safe resection, percutaneous radiation and in most cases, chemotherapy. Despite new radiation strategies and the development of oral alkylating substances (for example Temozolomide), the survival rate is still only approximately 15 months[13]. Although in former years the surgical role was controversial, current literature shows maximum safe surgical resection as a positive predictor for extended patient survival[14]. Microsurgical resection is currently optimized with the technical development of neuronavigation[15] containing functional datasets such as diffusion tensor imaging (DTI), functional magnetic resonance imaging (fMRI), magnetoence­phalography (MEG), magnetic resonance spectroscopy (MRS), or positron-emission-computed-tomography (PET). An early postoperative MRI with a contrast agent at the point of origin quantifies the tumor mass removal.

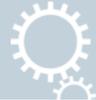

Then, the patient undergoes frequent MRI scans during the time of adjuvant therapy. Especially in case of a remnant tumor, the tumor volume has to be rigidly registered so a new tumor growth is not missed. For glioma segmentation in general (*WHO* grade I–IV), several MRI-based algorithms have been introduced in the literature.

A good overview of deterministic and statistical segmentation approaches is given by Angelini [16]. Most of these are region-based while the more recent ones are based on deformable models and include edge information. Segmentation based on outlier detection in T2-weighted MR data has been proposed by Prastawa *et al.*[17], whereby the image data is registered on a normal brain atlas to detect the abnormal tumor region. Sieg *et al.*[18] have introduced an approach to segment contrast-enhanced, intracranial tumors and anatomical structures of registered, multispectral MRI data. Using intensity-based pixel probabilities for tumor tissue, Droske *et al.*[19] have presented a deformable model, using a *level set*[20] formulation, to divide the MRI data into regions of similar image properties for tumor segmentation. An interactive method for segmentation of full-enhancing, ring-enhancing and non-enhancing tumors has been proposed by Letteboer *et al.*[21]. Clark *et al.*[22] introduced a knowledge-based automated segmentation method in order to partition glioblastomas. Gibbs *et al.*[23] introduced a combination of region growing and morphological edge detection for segmenting enhancing tumors in T1-weighted MRI data.

In the following, we describe studies that are more closely related to our contribution. For example, Song *et al.*[24] introduced a novel framework for automatic brain MRI tissue segmentation that overcomes inherent difficulties associated with this particular segmentation problem. They use a graph cut/atlas-based registration methodology that is iteratively optimized and incorporates probabilistic atlas priors and intensity-inhomogeneity correction for image segmentation. The usage of prior knowledge to guide the segmentation has been presented in a publication of Zhang *et al.*[25]. In a first step, they use the continuity among adjacent frames to generate a motion template according to the Displaced Frame Difference's (DFD) higher character and a color template is established by using k-means clustering. Afterwards (based upon the information derived from motion and the color templates), the segmentation image is defined as foreground, background and boundary regions. Finally, the segmentation problem is formulated as an energy minimization problem. Datteri *et al.*[26] proposed a combination of two segmentation methods: atlas based segmentation and spectral gradient graph cuts. To combine these two methods they first use the atlas-based segmentation method to segment the image. Then, they generate a third image used in the spectral gradient method as well as the source and sink points needed to initialize the graph cut algorithm.

This article is organized as follows. In Section 2, the experimental results are presented. Section 3 discusses our study and outlines areas for prospective tasks. Section 4 describes the details of the used material and the newly proposed approach. In part two of Section 4 (Calculation), a practical development from a theoretical basis is presented. Part three of Section 4 (Theory) extends the background of the contribution and lays the foundation for further work.

## Results

To implement the presented segmentation scheme, the *MeVisLab*-Platform (see http://www.mevislab.de) has been used and the algorithm has been implemented in C++ as an additional *MeVisLab*-module. Although the prototyping platform *MeVisLab* especially targets medical applications, it is possible to process images from other fields. Even when the graph was set up with a few hundred rays and hundreds of nodes were sampled along each ray, the overall segmentation (sending rays, graph construction and mincut computation) for our implementation took only a few seconds on an *Intel Core i5-750 CPU, 4x2.66 GHz, 8 GB RAM, Windows XP Professional x64 Version, Version 2003, Service Pack 2*.

For 2D evaluation, we used several synthetic and real images. Figure 1 shows a stone flounder (A). Stone flounders can blend into their environment by changing their color, and therefore it is difficult for the human eyes to detect them. In image B, the user-defined template of a stone flounder is shown that has been used for setting up the graph, and image C shows the nodes that have been generated with this template. In image D, the nodes are superimposed on the original image; the graph is twice as large as the template, therefore, the scaling of the stone flounder does not play a role (i.e. it is *scale invariant*), and the same template can be used for segmentation of smaller or larger stone flounders. Finally, image E presents the segmentation result.

In 3D, the algorithm has been evaluated on segmentation of brain tumors (glioblastoma multiforme and pituitary adenoma) from 60 clinical Magnetic Resonance Imaging datasets from an active neurosurgical practice (see also Supplementary Information). All brain tumors were somewhat spherically or elliptically shaped; therefore, we used the surface of a polyhedron to construct the graph. Segmentations performed by three neurosurgeons with several years of experience in the resection of brain tumors are considered the "gold standard" or "ground truth" against which we evaluate the results of our algorithm. A comparison yields an average Dice Similarity

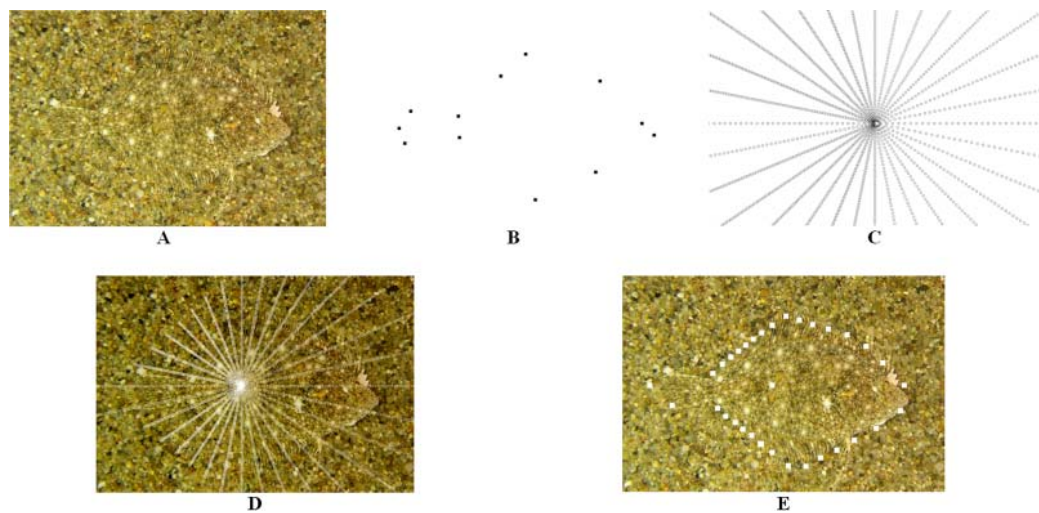

**Figure 1** | (A) Kareius bicoloratus (stone flounder). (B) User-defined template of the stone flounder. (C) Nodes set up with the template. (D) Nodes superimposed in the original image. (E) Segmentation result (white seed points).

 

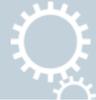

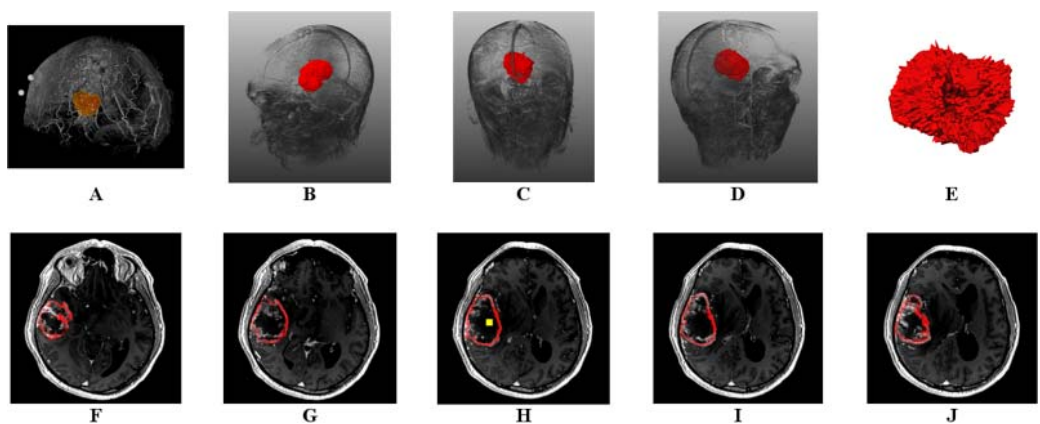

**Figure 2** | (A) 3D view of an automatically segmented tumor (brown). (B)–(D) Different 3D views of an automatically segmented tumor (red). (E) voxelized tumor mask. (F)–(J) Result of automatic tumor segmentation (DSC=81.33%). The yellow point (inside the tumor) in image H is the user-defined seed point. Manual segmentation performed by a neurological surgeon took 16 minutes for this dataset.

Coefficient (DSC)[27,28] of about 80%. The Dice Similarity Coefficient is a measure for spatial overlap of different segmentation results and is commonly used in medical imaging studies to quantify the degree of overlap between two segmented objects A and R, given by:

$$DSC = \frac{2 \cdot V(A \cap R)}{V(A) + V(R)} \quad (1)$$

The Dice Similarity Coefficient is the relative volume overlap between A and R, where A and R are the binary masks from the automatic A and the reference R segmentation. $V(\cdot)$ is the volume (in cm³) of voxels inside the binary mask, by means of counting the number of voxels, then multiplying with the voxel size.

Figure 2 shows some segmentation results for glioblastoma multiforme. In image A, a 3D view of an automatically segmented tumor (brown) is shown. The images B-D display different 3D views of an automatically segmented tumor (red), and the voxelized tumor mask is presented in image E. The images F-J show axial slices where the result of the automatic tumor segmentation is superimposed. The DSC for this segmentation is 81.33%, and the yellow point (inside the tumor) in image H is the user-defined seed point. Manual segmentation performed by a neurosurgeon took 16 minutes for this dataset. As shown in Figure 2, the segmentation works also with more elliptically shaped tumors. The algorithm only assumes that the object of interest is not extremely tubular, like vessels or the spinal cord. Also, the user-defined seed point does not have to be exactly in the center of the tumor, as shown in image H of Figure 2 (yellow). Even with a seed point that is located far from the center, the border of the tumor in Figure 2 (red) could still be recovered with a DSC of over 80% (note: the five axial slices F-J show only a small part of the tumor, the whole tumor was spread across 60 slices).

Image A of Figure 3 shows a graph (nodes and edges) constructed with a polyhedral surface to illustrate the dimensions of a typical graph used for pituitary adenoma segmentation. Image B of Figure 3 presents an axial slice of a pituitary adenoma with the segmented border superimposed and a zoomed-in view of the pituitary adenoma area for better illustration. The images C and D show different sagittal cross-sections with an automatically segmented pituitary adenoma (brown). Image E shows a 3D mask of an automatically segmented pituitary adenoma (red). Five axial slices with the superimposed border of the segmentation result (red) are presented in the images F-J where the user-defined seed point is located in image H (blue).

Table 1, Table 2 and Table 3 provide the results (minimum, maximum, mean $\mu$ and standard deviation $\sigma$) for all GBMs, pituitary adenomas and vertebrae that have been segmented with the presented algorithm and compared with manual slice-by-slice segmentations from the neurosurgeons. Table 1 provides results for fifty glioblastoma multiforme: volume of tumor (cm³), number of voxels and Dice Similarity Coefficient. In Table 2, the results for ten pituitary adenomas are presented: volume of tumor (cm³), number of

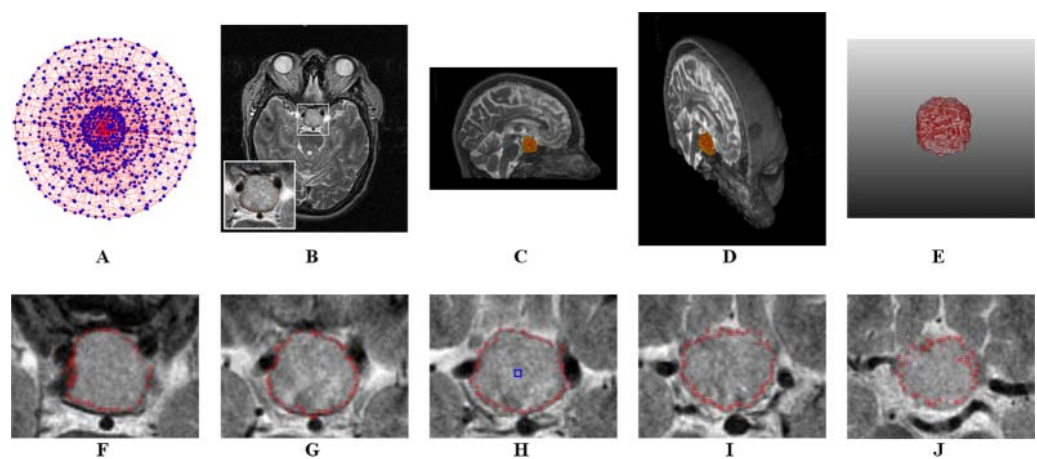

**Figure 3** | (A) Graph (nodes and edges) constructed with a polyhedron surface. (B) Axial slice of a pituitary adenoma. (C) (D) Different views of sagittal slices with an automatic segmented pituitary adenoma. (E) 3D mask of an automatically segmented pituitary adenoma. (F)–(J) Segmentation results for a pituitary adenoma dataset. (H) user-defined seed point (blue).

                                                                                    

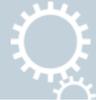

**Table 1 | Summary of results: min., max., mean $\mu$ and standard deviation $\sigma$ for fifty glioblastoma multiforme (GBM)**

|  | Tumor Volume (cm³) | | Voxel Number | | |
|  | manual | automatic | manual | automatic | DSC (%) |
|---|---|---|---|---|---|
| min | 0.47 | 0.46 | 524 | 783 | 46.33 |
| max | 119.28 | 102.98 | 1024615 | 884553 | 93.82 |
| $\mu \pm \sigma$ | 23.66±24.89 | 21.02±22.90 | 145305.54 | 137687.24 | 80.37±8.93 |

voxels, Dice Similarity Coefficient and the manual segmentation times (minutes). Finally, Table 3 shows the results for vertebrae segmentation: volume of vertebra (cm³), number of voxels, and Dice Similarity Coefficient. For a direct comparison and discussion of our method with other methods from the literature we refer the reader to previous publications[29,30]. In the first contribution[29], the results of vertebral segmentation – based on a rectangle shape – are compared with an interactive multi-label N-D image segmentation method called *GrowCut* from Vezhnevets and Konouchine[31]. In the other contribution[30], our method – based on a spherical template – is directly compared with a balloon inflation approach[32] for *WHO* grade IV glioma segmentation. Finally, we refer the reader to an additional publication where our template-based approach has been used to segment the bladder for MR-guided brachytherapy for gynecologic malignancies[33].

## Discussion

In this contribution, we have presented a template-based segmentation scheme for 2D and 3D objects. To the best of our knowledge, this is the first approach where the nodes of a graph-based algorithm have been arranged according to a predefined template in a non-uniform and a non-equidistant manner on an image. Using this new type of segmentation algorithm, it is possible to reconstruct missing arcs and kinks in an object. In addition, the presented method is scale invariant. Experimental results for several 2D and 3D images based on 60 Magnetic Resonance Imaging datasets consisting of two types of brain tumors, glioblastoma multiforme and pituitary adenoma, indicate that the proposed algorithm requires less computing time and gives results comparable to human experts using a simple cost function. The presented work is a generalization of recent work by the authors[29,34] to arbitrary user-defined shapes in 2D and 3D. In previous work[34], a system for volumetric analysis of cerebral pathologies was introduced that used a sphere template for the segmentation process and therefore was limited to spherically-shaped objects. In[29] a rectangle-based segmentation algorithm for vertebrae MR images was introduced. As stated in the background section of the Introduction, there are proposed approaches from Song et al.[24], Zhang et al.[25] and Datteri et al.[26] that use prior knowledge like motion and color templates and shape information in graph based approaches. However, these approaches do not distribute the graphs nodes non-uniformly and non-equidistantly on the image. Instead they work on a regular grid, and compensate by adding complexity to the objective function. In summary, the achieved research highlights of the presented work are:

- A template-based segmentation paradigm for 2D and 3D objects
- Nodes are arranged according to a predefined template

- The approach represents a new type of graph-based algorithms
- It is possible to reconstruct missing arcs and kinks in an object
- The method is scale invariant

In our experience, and that of most applied researchers, automatic segmentation methods are served well by companion editing tools that can be used to efficiently "clean up" the results when needed. Therefore, we developed a manual refinement method that takes advantage of the basic design of graph-based image segmentation algorithms[35]. The manual refinement method can also be used for any graph-based image segmentation algorithms and therefore also for the template-based segmentation scheme. For twelve GBM cases, the Wilcoxon signed-rank test[36,37] verified a significant improvement (p=0.016) for our manual refinement method for a significance level of 0.05. However, the results presented in this study are not based on the use of manual refinement after the initial segmentations.

There are several areas of future work. For example, the cost function for the weights can be customized. Another possibility is to increase the sampling rate (for the nodes) near an object's border, because – with an equidistant sampling rate (along the rays) – there are more nodes near the user-defined seed point and less nodes going farther out. Moreover, the user-defined seed point position that is located inside the object is also an issue that can be analyzed in the future, e.g. for the stone flounder, the seed point has to be chosen carefully. One option to improve the presented algorithm is to perform the segmentation iteratively: After segmentation has been performed, the center of gravity of the segmentation can be used as a new seed point for a segmentation and so on. This may lead to increased robustness with respect to the initialization.

Finally, we point the interested reader to publications from Sharon et al.[38] and Corso et al.[39] that are based on algebraic multigrid methods and graph cuts (normalized cuts) in which they introduced methods that adaptively build a graph and approximate cuts at varying resolutions and scales. A combination of their proposed method with our approach would result in an interesting template-based graph at every level.

## Methods

The proposed segmentation scheme starts by setting up a directed graph from a user-defined seed point that is located inside the object to be segmented. To set up the graph, points are sampled along rays cast through the contour (2D) or surface (3D) of an object template. The sampled points are the nodes $n \in V$ of the graph $G(V, E)$ and $e \in E$ is the corresponding set of edges. There are edges between the nodes and edges that connect the nodes to a source $s$ and a sink $t$ to allow the computation of an s-t cut (note: the source and the sink $s, t \in V$ are virtual nodes). Similar to the notation introduced by Li et al.[4], the arcs $<v_i,v_j> \in E$ of the graph G connect two nodes $v_i,v_j$. There are two types of ∞-weighted arcs: p-arcs $A_p$ and r-arcs $A_r$ (P is the number of sampled points along one ray r=(0,...,P-1) and R is the number of rays cast to the contour or surface of an object template $r=(0,...,R-1)$), where $V(x_n,y_n)$ is a neighbor of V(x,y) – in other words V(x$_n$,y$_n$) and V(x,y) belong to two adjacent rays. For a surface in 3D, the principle is the same, except that there is an additional dimension for a node (V(x,y,z)):

$$A_p = \{\langle V(x, y), V(x, y-1)\rangle | y > 0\}$$
$$A_r = \{\langle V(x, y), V(x_n, \max(0, y - \Delta_r))\rangle\} \tag{2}$$

$$A_p = \{\langle V(x, y, z), V(x, y, z-1)\rangle | z > 0\}$$
$$A_r = \{\langle V(x, y, z), V(x_n, y_n, \max(0, z - \Delta_r))\rangle\} \tag{3}$$

**Table 2 | Summary of results: min., max., mean $\mu$ and standard deviation $\sigma$ for ten pituitary adenomas**

|  | PA Volume (cm³) | | Voxel Number | | | Manual seg. time (minutes) |
|  | manual | automatic | manual | automatic | DSC (%) | |
|---|---|---|---|---|---|---|
| min | 0.84 | 1.18 | 4492 | 3461 | 71.07 | 3 |
| max | 15.57 | 14.94 | 106151 | 101902 | 84.67 | 5 |
| $\mu \pm \sigma$ | 6.30±4.07 | 6.22±4.08 | 47462.7 | 47700.6 | 77.49±4.52 | 3.91±0.54 |

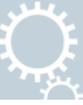

**Table 3 | Summary of results: min., max., mean $\mu$ and standard deviation $\sigma$ for nine vertebrae.**

|  | Vertebrae Volume (cm³) | | Voxel Number | | |
|---|---|---|---|---|---|
|  | manual | automatic | manual | automatic | DSC (%) |
| min | 0.25 | 0.24 | 1015 | 995 | 87.37 |
| max | 0.51 | 0.49 | 2091 | 2010 | 94.93 |
| $\mu \pm \sigma$ | 0.42±0.072 | 0.40±0.073 | 1722 | 1656 | 90.97±2.2 |

The arcs between two nodes along a ray A$_p$ ensure that all nodes below the contour or surface in the graph are included to form a closed set (correspondingly, the interior of the object is separated from the exterior in the data). This principle is shown in Figure 4 on the left side (A) for two rays of a circular template. The arcs A$_r$ between the nodes of different rays constrain the set of possible segmentations and enforce smoothness via the regularization parameter $\Delta_r$, that controls the stiffness of the surface. A delta value of zero ensures that the segmentation result has exactly the form of the predefined template – and the position of the template depends on the best fit to the gray levels or appearance of the image. The weights w(x,y) for every edge between $v \in V$ and the sink or source are assigned in the following manner: weights are set to c(x,y) if z is zero; otherwise they are set to c(x,y)-c(x,y-1), where c(x,y) is the absolute value of the difference between an average texture value of the desired object and the texture value of the pixel at position (x,y) – for a detailed description see[41–43]. The average texture value in this case, or the cost function in the general case, as well as the weights, critically influence the segmentation result. Based on the assumption that the user-defined seed point is inside the object, the average gray value can be estimated automatically. Therefore, we integrate over a small square S (2D) or cube C (3D) of dimension $d$ centered on the user-defined seed point (s$_x$, s$_y$) in the 2D case and (s$_x$, s$_y$, s$_z$) in the 3D case:

$$\int_{-d/2}^{d/2} \int_{-d/2}^{d/2} S(s_x + x, s_y + y)\,dxdy \tag{4}$$

$$\int_{-d/2}^{d/2} \int_{-d/2}^{d/2} \int_{-d/2}^{d/2} C(s_x + x, s_y + y, s_z + z)\,dxdydz \tag{5}$$

The principle underlying the graph construction for a square is shown in Figure 5. Image A of Figure 5 shows the corners of a square template that are used to set up the graph. Image B shows the nodes that have been sampled along the rays that have been sent through the template's surface. Note that the distances between the nodes of one ray correlate with the distances between the template's center point (or for a later segmentation, the user-defined seed point) and the template surface. In other words, for every ray we have the same number of nodes between the center point and the object's border, but the length is different. In images C, D and E, different ∞-weighted arcs are shown: C: the p-arcs A$_p$ along the single rays, D: the r-arcs A$_r$ between rays with a delta value of A$_r$=0. E: same as D only with a delta value of A$_r$=1.

For the 3D case, both, the automatic segmentation method and a manual slice-by-slice segmentation performed by a domain expert (for a later evaluation of the automatic segmentation result) are post-processed in an identical manner. The resulting contours (given as point clouds) of the object's boundaries are triangulated to get a closed surface. This closed surface is used to generate a solid 3D mask (representing the segmented object), which is achieved by voxelization of the triangulated mesh[44].

The overall workflow of the introduced segmentation scheme is presented in Figure 6. In the upper row, a 2D square template is used for vertebral segmentation. In the lower row, a 3D sphere template is used to segment a GBM.

**Calculation.** Setting up the nodes of the graph with the user-defined template is the step that requires the most ingenuity in the proposed algorithm. Generating the arcs between the nodes and the source and the sink node is straightforward: there are the ∞-weighted arcs that depend on the geometry (intracolumn arcs) and the delta value (intercolumn arcs) used for the graph, and there are arcs that connect the nodes to the source $s$ and the sink $t$. These arcs depend on the gray values of the nodes they connect – or rather they depend on the gray value difference to an adjacent node. To integrate the user-defined template into the construction of the graph, we need the coordinates in 2D or 3D describing the object that we want to segment (e.g. for a square the edges of the square, see Figure 5 A). Using these coordinates, the center of gravity of the object is calculated, and the object is normalized with the maximum diameter, or rather with the coordinate that has the maximum distance to the center of gravity. After the user defines a seed point in the image (2D) or volume (3D), the normalized object is constructed with its center of gravity point located at the user-defined seed point. Then, rays are drawn radially (2D) or spherically (3D) out from the seed point through the contour (2D) or surface (3D) of the normalized object. To calculate the intersection points of the rays with the object, its contour (in 2D) or surface (in 3D) has to be closed. In our implementation, we assume that the user provides the object's contour as 2D coordinates ordered in the clockwise direction, and we connect the points one after the other and finally connect the last point with the first point to get a closed 2D contour. To get a closed surface of the 3D objects, the object is triangulated[45]. However, this is not necessary for a spherical or elliptical segmentation where a sphere is used as a template. Thus, the computing time for the triangulation and the following ray-triangle intersection calculation can be avoided by using the surface points of a polyhedron. Their surface coordinates already provide the locations where the rays have to be sent through.

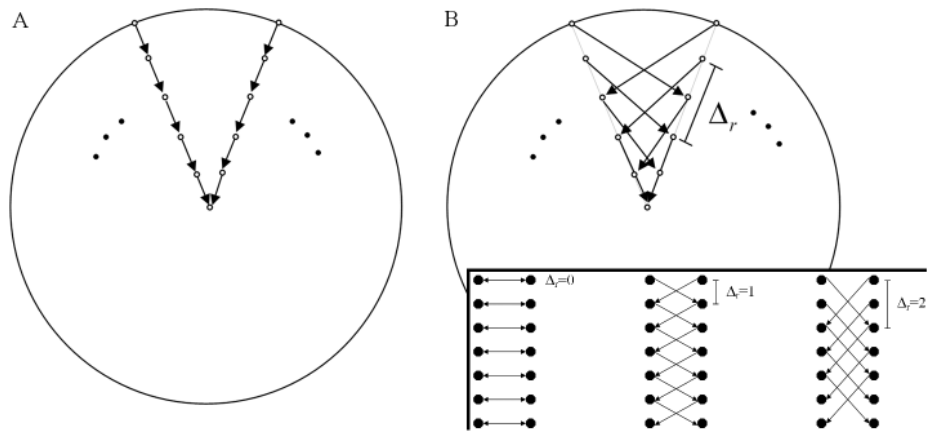

**Figure 4 | The two different types of arcs for a graph that is used to segment circular shaped objects: A$_p$ arcs (A) and A$_r$ arcs (B). Lower right: Intercolumn edges for: $\Delta_r$=0 (left), $\Delta_r$=1 (middle) and $\Delta_r$=2 (right).**

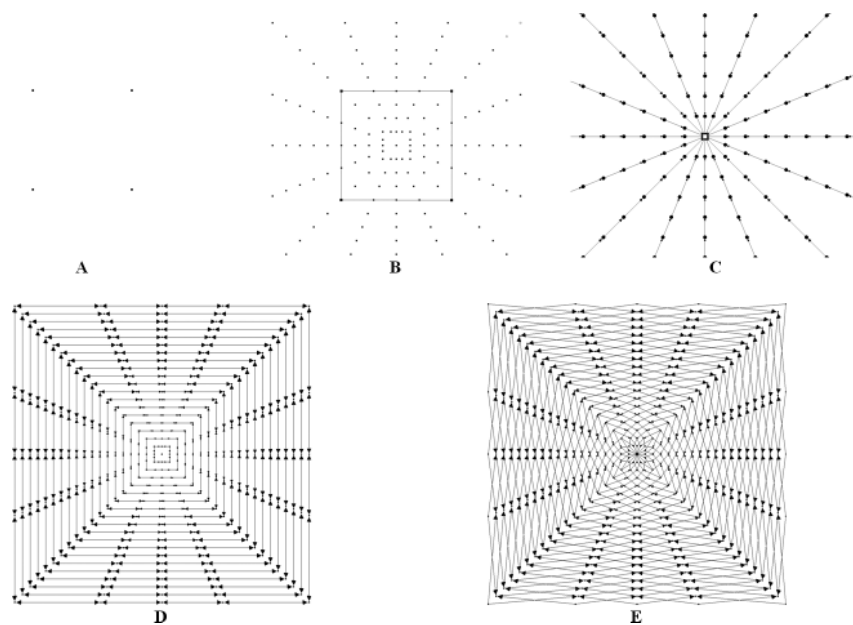

**Figure 5** | (A) Square template given by its corners. (B) Nodes set up with the template. (C) p-arcs $A_p$ along the rays. (D) r-arcs $A_r$ between rays ($\Delta_r = 0$). (E) r-arcs $A_r$ between rays ($\Delta_r = 1$).

The intersection point of a ray with the object provides the distance between the nodes for this ray, because all rays have the same number of nodes from the center of gravity point to the intersection with the contour or surface. For intersections that are located closer to the center of gravity point we get smaller distances, and for intersections that are located farther away from the center of gravity point we get larger distances. Calculating the intersection of a ray with a 2D object is straightforward, since it is simply a line-line intersection. One line is the actual ray and the other line is one straight line between two adjacent points of the predefined template. Since triangulated objects are used for 3D segmentation, ray-triangle intersections for the 3D template segmentation have to be calculated. To implement ray-triangle intersections, there are several fast algorithms, such as the algorithms proposed by Möller and Trumbore[46] and by Badouel[47]. Given that the calculations of the ray-triangle intersections require the largest fraction of computing power, a GPU realization of these calculations[48] is used for objects with complex shapes.

**Theory.** The procedure of setting up the nodes of the graph based on the template biases the cut towards a particular shape, and the delta value $\Delta_r$ is a regularizer that influences the variations of the results. In other words, the delta value $\Delta_s$ specifies how much the segmentation results are allowed to deviate from the user-defined template. For example, a delta value of zero ($\Delta_s = 0$) forces the segmentation result to have the exact shape of the template, which is optimal for problems where the shape but not the scale of the object is known. As is the case with regularizers in general, the delta value has to be chosen carefully corresponding to the segmentation problem. On the one hand, the segmentation results should not be too "stiff" with respect to the template, such that the algorithm is not flexible enough to handle some variations of the object and miss them during the segmentation process. On the other hand, if the delta value is too large, one risks obtaining results with shapes that do not correspond to the predefined template anymore. We studied this tradeoff for vertebral segmentation with a square template for different delta values[29]. Principal component analysis (PCA), also known as Karhunen-Loeve transform[49] can potentially be used as an interesting mechanism for incorporating additional domain knowledge about the shape of the target object into our algorithm, and also better informing the selection of the value for the delta regularizing parameter. This concept of using PCA for characterizing shapes is well developed in Active Shape Models (ASM)[50] and the shape model formulation of the Active Appearance Models (AAM)[51]. AAMs model the variability of shapes within an object class by removing variation introduced by rotational, translational and scaling effects from the training shapes, and all shapes need to be aligned to each other with respect to the mentioned transformations before a statistical analysis can be done. After the principal modes of variation (and corresponding eigenvalues) are computed from training data, legal shape instance $s$ contained in the distribution derived from the training set can be generated from the model by deforming the *mean shape $\bar{s}$* by a linear combination of eigenvectors. Thus, the shape model is described by

$$s = \bar{s} + \Phi_s b_s \tag{6}$$

where $b_s$ is a vector containing the model parameters weighting the contribution of each eigenvector to the deformation, and $\Phi_s$ are the eigenvectors. To incorporate this shape model formulation into the template-based approach introduced in this contribution, first, a *mean shape* of the target object can be computed from several registered, manual segmentations using standard PCA. This *mean shape $\bar{s}$* can then be input as the template or the distribution of the graph's nodes for our method. The next step is to establish a relationship between the variations of the object from the mean shape (as are obtained in standard PCA), and the delta regularizer of our algorithm. We believe that a reasonable scalar estimate of the delta value $\Delta_r$ can be computed proportional to the quantity max($\Phi_s b_s$).

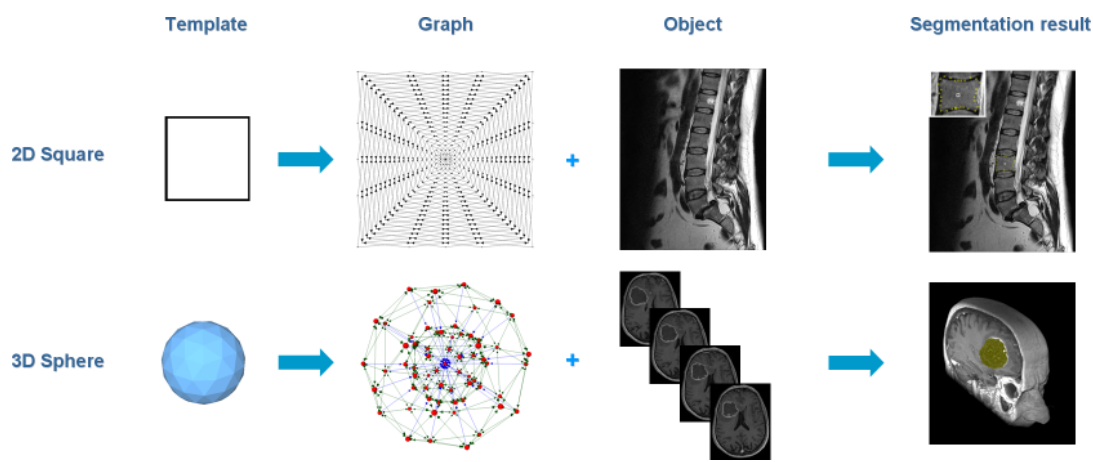

**Figure 6** | **Principle workflow of the presented segmentation scheme in 2D and 3D.** In 2D a square template is used to segment a vertebra. In 3D a sphere template is used to segment a glioblastoma multiforme (GBM).

## Acknowledgements

First of all, the authors would like to thank the physicians Dr. med. Barbara Carl, Thomas Dukatz, Christoph Kappus and Dr. med. Daniela Kuhnt from the neurosurgery department of the university hospital in Marburg for performing the manual slice-by-slice segmentations of the medical images, therefore, providing the ground truth for the evaluation. In addition, the authors would like to thank Brandon Greene from the Institute for Medical Biometry and Epidemiology (IMBE) of the University of Marburg for helping with the statistics. Moreover, the authors would like to thank Fraunhofer MeVis in Bremen, Germany, for their collaboration and especially Horst K. Hahn for his support.
This project was supported by the *National Center for Research Resources (P41RR019703)* and the *National Institute of Biomedical Imaging and Bioengineering (P41EB015898, U54EB005149, R03EB013792)* of the *National Institutes of Health (NIH)*. Its contents are solely the responsibility of the authors and do not necessarily represent the official views of the *NIH*.

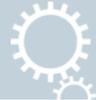

## Author contribution

Conceived and designed the experiments: JE. Performed the experiments: JE. Analyzed the data: JE CN. Contributed reagents/materials/analysis tools: CN BF JE TK. Wrote the paper: JE TK BF.

## Additional information

**Competing financial interest:** All authors in this paper have no competing financial interests.